\documentclass[10pt,twocolumn,letterpaper]{article}
\usepackage{cvpr}
\usepackage{times}
\usepackage{epsfig}
\usepackage{graphicx}
\usepackage{amsmath}
\usepackage{amssymb}
\usepackage{gensymb}
\usepackage{pbox}
\usepackage{color}
\usepackage{setspace}

\makeatletter
\def\blfootnote{\xdef\@thefnmark{}\@footnotetext}
\makeatother

\usepackage[pagebackref=true,breaklinks=true,letterpaper=true,colorlinks,bookmarks=false]{hyperref}

\cvprfinalcopy 

\ifcvprfinal\fi
\begin{document}

\title{Normal Assisted Stereo Depth Estimation}

\author{ Uday Kusupati$^{1*}$ \hspace{.8cm} Shuo Cheng$^2$ \hspace{.8cm} Rui Chen$^{3*}$ \hspace{0.8cm}
Hao Su$^2$ \vspace{.2cm}\\ 
\vspace{-0.0cm}
$^1$The University of Texas at Austin \hspace{.8cm} $^2$University of California San Diego \vspace{.2cm}\\ 
$^3$Tsinghua University\\
{\tt\small uday@cs.utexas.edu, scheng@eng.ucsd.edu, chenr17@mails.tsinghua.edu.cn, haosu@eng.ucsd.edu}
}

\maketitle

\begin{abstract}
   Accurate stereo depth estimation plays a critical role in various 3D tasks in both indoor and outdoor environments. Recently, learning-based multi-view stereo methods have demonstrated competitive performance with limited number of views. However, in challenging scenarios, especially when building cross-view correspondences is hard, these methods still cannot produce satisfying results. In this paper, we study how to leverage a normal estimation model and the predicted normal maps to improve the depth quality. We couple the learning of a multi-view normal estimation module and a multi-view depth estimation module. In addition, we propose a novel consistency loss to train an independent consistency module that refines the depths from depth/normal pairs. We find that the joint learning can improve both the prediction of normal and depth, and the accuracy \& smoothness can be further improved by enforcing the consistency. Experiments on MVS, SUN3D, RGBD and Scenes11 demonstrate the effectiveness of our method and state-of-the-art performance.
\end{abstract}





\section{Introduction}
\begin{figure}[t]
\setlength{\belowcaptionskip}{-1.0cm}
\begin{center}
  \includegraphics[width=\linewidth]{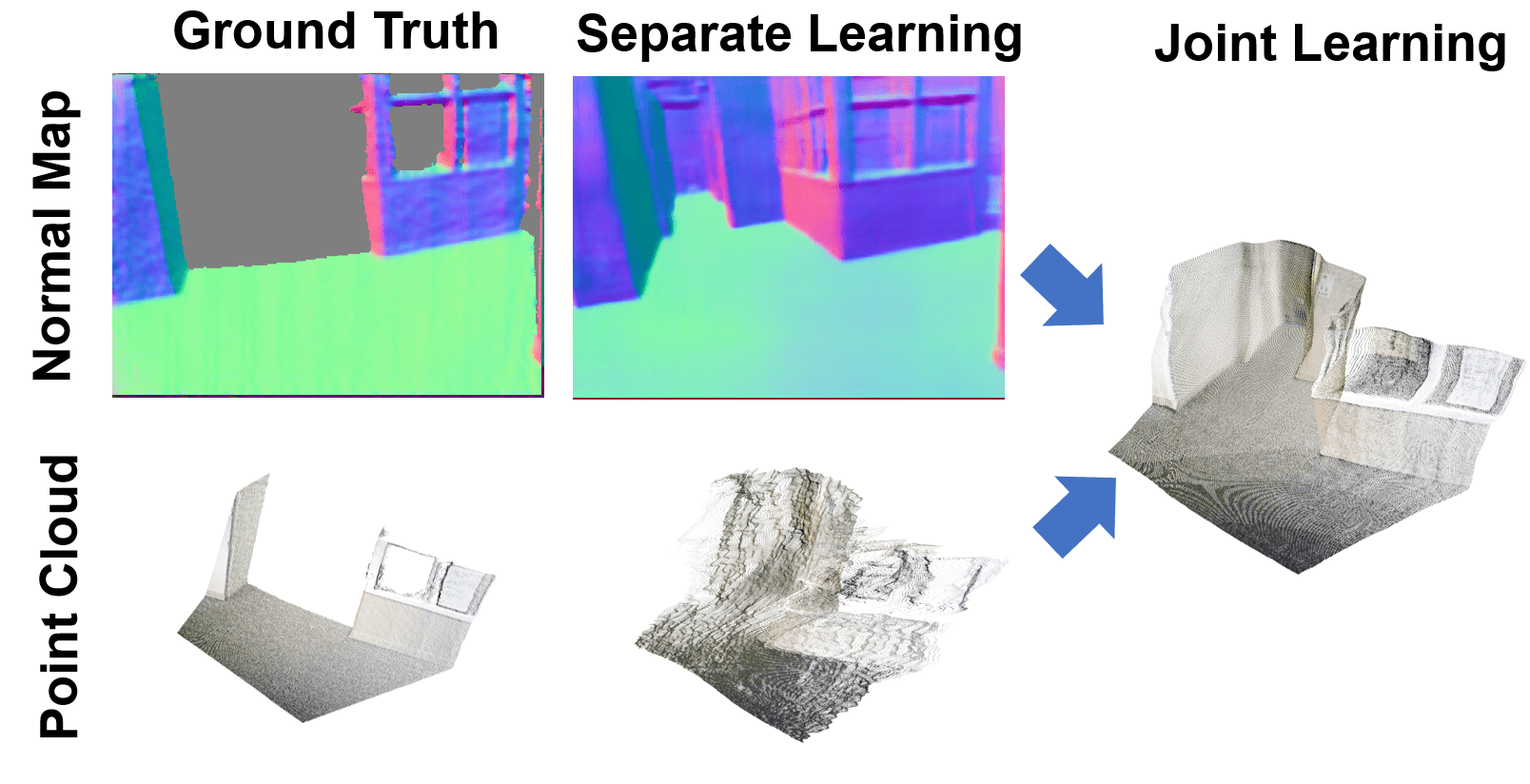}
\end{center}
  \caption{Illustration of results of separate learning and joint learning of depth and normal. While the normal prediction is smooth and accurate, existing state-of-the-art stereo depth prediction result is noisy. Our method improves the prediction quality significantly by joint learning of depth and normal and enforcing consistency. \textbf{Color format. Best viewed on screen.}
  }
\label{fig:teaser}
\end{figure}

\blfootnote{$^*$Work done while visiting University of California San Diego}
Multi-view stereo (MVS) is one of the most fundamental problems in computer vision and has been studied over decades. Recently, learning-based MVS methods have witnessed significant improvement against their traditional counterparts~\cite{yao2018mvsnet, DBLP:conf/iclr/ImJLK19, yao2019recurrent, ChenPMVSNet2019ICCV}. In general, these methods formulate the task as an optimization problem, where the target is to minimize the overall summation of pixel-wise depth discrepancy. However, the lack of geometric constraints leads to bumpy depth prediction especially in areas with low texture or that are textureless as shown in Fig.~\ref{fig:teaser}. Compared with depth that is a property of the global geometry, surface normal represents a more local geometric property and can be inferred more easily from visual appearance. For instance, it is much easier for humans to estimate whether a wall is flat or not than the absolute depth.  Fig.~\ref{fig:teaser} shows an example where learning-based MVS methods perform poorly on depth estimation but significantly better on normal prediction. 

Attempts have been made to incorporate the normal based geometric constraints into the optimization to improve the monocular depth prediction~\cite{yin2018geonet, DBLP:conf/cvpr/ZhangSYSLJF17}.
One simple form of enforcing a consistency constraint between depth and normal is to enforce orthogonality between the predicted normal and the tangent directions computed from the predicted depths at every point. However, for usage as regularizing loss function during training, we find that a naive consistency method in the world coordinate space is a very soft constraint as there are many sub-optimal depth solutions that are consistent with a given normal.
Optimizing depths to be consistent with the normal as a post processing\cite{DBLP:conf/cvpr/ZhangSYSLJF17} ensures local consistency; however, not only is this an expensive step during inference time, but also the post-processed result may lose grounding from the input images. Therefore, we strive to propose a new formulation of depth-normal consistency that can improve the training process. Our consistency is defined in the pixel coordinate space and we show that our formulation is better than the simple consistency along with better performance than previous methods to make the geometry consistent. This constraint is independent of the multi-view formulation and can be used to enforce consistency on any pair of depth and normal even in the single view setting. To this end, our contributions are mainly in the following aspects:

First, we propose a novel cost-volume-based multi-view surface normal prediction network (NNet). By constructing a 3D cost volume by plane sweeping and accumulating the multi-view image information to different planes through projection, our NNet can learn to infer the normal accurately using the image information at the correct depth. The construction of a cost volume with image features from multiple views contains the information of available features in addition to enforcing additional constraints on the correspondences and thus the depths of each point. We show that the cost volume is a better structural representation that facilitates better learning on the image features for estimating the underlying surface normal. While in single image setting, the network tends to overfit the texture and color and demonstrates worse generalizability, we show that our method of normal estimation generalizes better due to learning on a better abstraction than single view images.

Further, we demonstrate that learning a normal estimation model on the cost volume jointly with the depth estimation pipeline facilitates both tasks. Both traditional and learning-based stereo methods suffer from the noisy nature of the cost volume. The problem is significant in textureless surfaces when the image feature based matching doesn't offer enough cues. We show that enforce the network to predict accurate normal maps from the cost volume results in regularizing the cost volume representation, and thereby assists in producing better depth estimation. Experiments on MVS, SUN3D, RGBD, Scenes11, and Scene Flow datasets demonstrate that our method achieves state-of-the-art performance.




\section{Related Work}
In this section we review the literature relevant to our work concerned with stereo depth estimation, normal estimation and multi-task learning for multi-view geometry.

\textbf{Classical Stereo Matching.}
A stereo algorithm typically consists of the following steps: matching cost calculation, matching cost aggregation and disparity calculation. As the pixel representation plays a critical role in the process, previous literature have exploited a variety of representations, from the simplest RGB colors to hand-craft feature descriptors~\cite{yoo2009fast, yang2008stereo, tola2009daisy, lowe1999object, bay2006surf}. Together with post-processing techniques like Markov random fields~\cite{szeliski2006comparative} and semi-global matching~\cite{hirschmuller2007stereo}, these methods can work well on relative simple scenarios.

\textbf{Learning-based Stereo.}
To deal with more complex real world scenes, recently researchers leverage CNNs to extract pixel-wise features and match correspondences~\cite{ji2017surfacenet, zbontar2016stereo, kendall2017end, Chang_2018_CVPR, Liang_2018_CVPR, khamis2018stereonet, Hartmann2017LearnedMS}. The learned representation shows more robustness to low-texture regions and various lightings~\cite{DeepMVS, yao2018mvsnet, yao2019recurrent, ChenPMVSNet2019ICCV, Luo_2019_ICCV}. Rather than directly estimating depth from image pairs as done in many previous deep learning methods, some approaches also tried to incorporate semantic cues and context information in the cost aggregation process~\cite{yang2018segstereo,cherabier2018learning, im2019dpsnet} and achieved positive results. While other geometry information such as normal and boundary~\cite{zhang2017robust, galliani2016gipuma, kazhdan2006poisson} are widely utilized in traditional methods for further improving the reconstruction accuracy, it is non-trivial to explicit enforce the geometry constraints in learning-based approaches~\cite{furukawa2015multi}. To the best of our knowledge, this is the first work that tries to solve depth and normal estimation in multi-view scenario in a joint learning fashion.

\textbf{Surface Normal Estimation.}
Surface normal is an important geometry information for 3D scene understanding. Recently, several data-driven methods have achieved promising results~\cite{Eigen_2015_ICCV, bansal2016marr, fouhey2014unfolding, wang2015designing, boulch2016deep, zeng2019deep}. 
While these methods learn the image level features and textures to address normal prediction in a single image setting, we propose a multi-view method that generalizes better and reduces the learning complexity of the task. 

\textbf{Joint Learning of Normal and Depth.}
With deep learning, numerous methods have been proposed for joint learning of normal and depth~\cite{qi2018geonet, hane2015direction, zhan2019self, dharmasiri2017joint, yin2019enforcing, zhang2018deep, Eigen_2015_ICCV}. 
Even though these methods achieved some progress, all these methods focus on single image scenario, while there are still few works exploring joint estimation of normal and depth in multi-view setting. Gallup~\etal~\cite{DBLP:conf/cvpr/GallupFMYP07} estimated candidate plane directions for warping during plane sweep stereo and further enforce an integrability penalty between the predicted depths and the candidate plane for better performance on slanted surfaces, however, the geometry constraints are only applied in a post processing or optimization step (e.g., energy model or graph cut). The lack of an end-to-end learning mechanism makes their methods easier to get stuck in sub-optimal solutions. In this work, our experiments demonstrate that with careful design, the joint learning of normal and depth is favorable for both sides, as the geometry information is easier to be captured. Benefited from the learned powerful representations, our approach achieves competitive results compared with previous methods.








\begin{figure*}[ht]
\setlength{\abovecaptionskip}{0pt} 
\begin{center}
\includegraphics[width=0.85\linewidth]{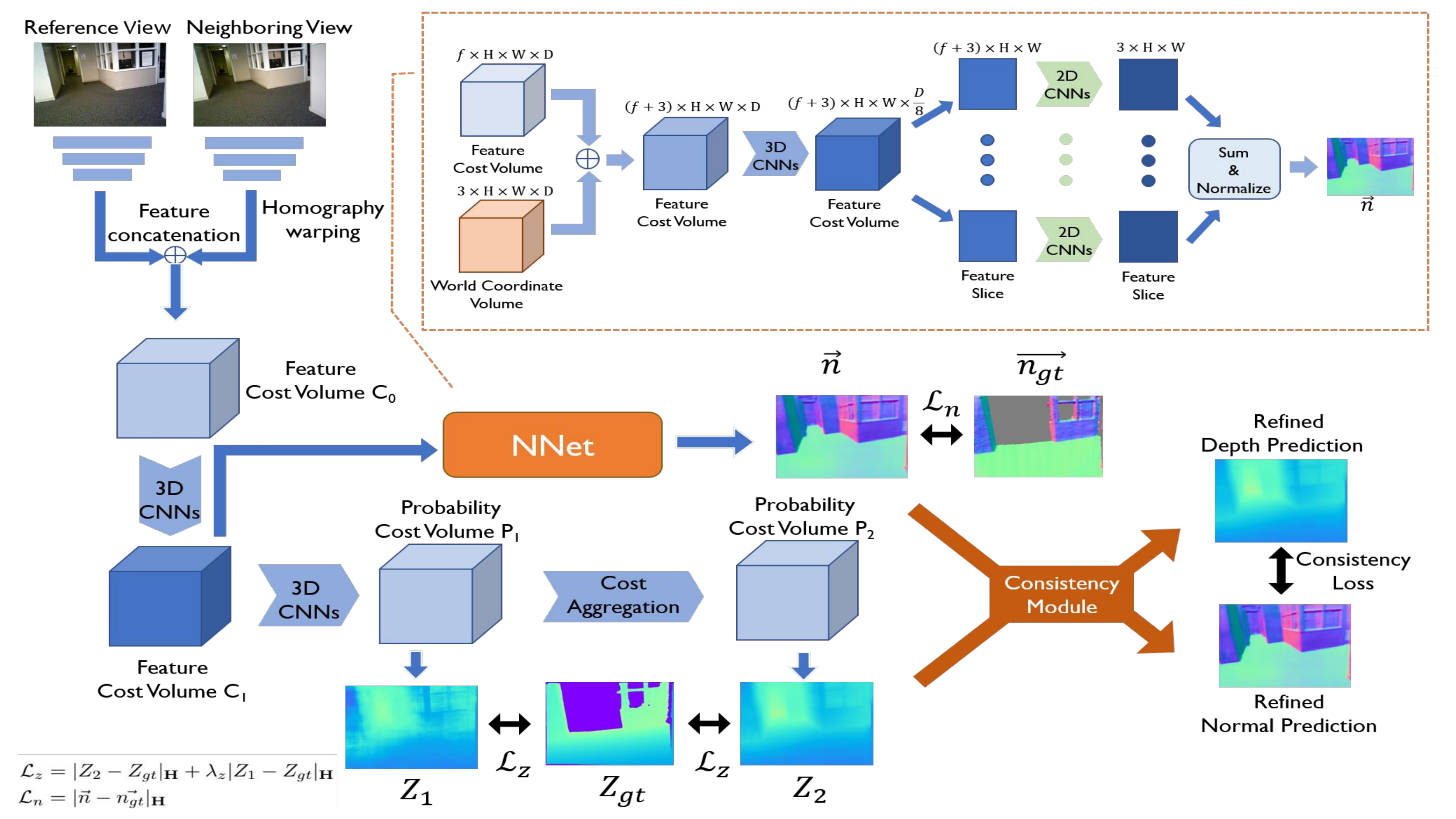}
\end{center}
  \caption{\textbf{Illustration of the pipeline of our method.} We first extract deep image features from viewed images and build a feature cost volume by using feature wrapping. The depth and normal are jointly learned in a supervised fashion. Further we use our proposed consistency module to refine the depth and apply a consistency loss.}
\label{fig:pipeline}
\end{figure*}

\section{Approach}
We propose an end-to-end pipeline for multi-view depth and normal estimation as shown in Fig.~\ref{fig:pipeline}. The entire pipeline can be viewed as two modules. The first module consists of joint estimation of depth and normal maps from the cost volume built from multi-view image features. The subsequent module refines the predicted depth by enforcing consistency between the predicted depth and normal maps using the proposed consistency loss. In the first module, joint prediction of normal from the cost volume implicitly improves the learned model for depth estimation.  The second module is explicitly trained to refine the estimates by enforcing consistency.

\subsection{Learning based Plane Sweep Stereo}

First, we describe our depth prediction module. In terms of the type of target, current learning-based stereo methods can be divided into two categories: single object reconstruction~\cite{DBLP:conf/eccv/YaoLLFQ18, ChenPMVSNet2019ICCV} and scene reconstruction~\cite{DBLP:conf/iclr/ImJLK19, DeepMVS}. Compared with single object reconstruction, scene reconstruction, where multiple objects are included, requires larger receptive field for the network to better infer the context information. Because this work also aims at scene reconstruction, we take DPSNet~\cite{DBLP:conf/iclr/ImJLK19}, a state-of-the-art scene reconstruction method, as our depth prediction module.

The inputs to the network are a reference image $\mathbf{I_1}$ and a neighboring view image $\mathbf{I_2}$ of the same scene along with the intrinsic camera parameters and the extrinsic transformation between the two views. We first extract deep image features using a spatial pyramid pooling module. Then a cost volume is built by plane sweeping and 3D CNNs are applied on it. Multiple cost volumes can be built and averaged when multiple views are present. Further context-aware cost aggregation~\cite{DBLP:conf/iclr/ImJLK19} is used to regularize the noisy cost volume. The final depth is regressed using soft argmin~\cite{DBLP:journals/corr/KendallMDHKBB17} from the final cost volume.

\subsection{Cost Volume based Surface Normal Estimation}\label{ssec: nnet}
In this section, we describe the network architecture of cost volume based surface normal estimation (Fig.~\ref{fig:pipeline}: NNet). The cost volume contains all the spatial information in the scene as well as image features in it. The probability volume models a depth distribution across candidate planes for each pixel. In the limiting case of infinite candidate planes, an accurately estimated probability volume turns out to be the implicit function representation of the underlying surface~\ie takes value 1 where a point on the surface exists and 0 everywhere else. This motivates us to use the cost volume $\mathbf{C_1}$ which also contains the image-level features to estimate the surface normal map ${\vec{\mathbf{n}}}$ of the underlying scene. 

Given the cost volume $\mathbf{C_1}$ we concatenate the world coordinates of every voxel to its feature. Then, we use three layers of 2-strided convolution along the depth dimension to reduce the size of this input to $\mathbf{((f+3){\times}H{\times}W{\times}D/8)}$ and call this $C_n$. Consider a fronto-parallel slice $S_i$ of size $\mathbf{((f+3){\times}H{\times}W)}$ in $C_n$. We pass each slice through a normal-estimation network (\textbf{NNet}). NNet contains 7 layers of 2D convolutions of $\mathbf{3\times3}$ with increasing receptive field as the layers go deep using dilated convolutions (1, 2, 4, 6, 8, 1, 1). We add the output of all slices and normalize the sum to obtain the estimate of the normal map.
\begin{equation}
\vec{n} = \frac{\sum_{i=1}^{D/8}\space \mathbf{NNet}(S_i)}{\big|\big|\sum_{i=1}^{D/8} \space\mathbf{NNet}(S_i)\big|\big|_2}
\end{equation}

We explain the intuition behind this choice as follows. Each slice contains information corresponding to the patch match similarity of each pixel in all the views conditioned on the hallucinated depths in the receptive field of the current slice. In addition, due to the strided 3D convolutions, the slice features accumulate information about features of a group of neighboring planes. The positional information of each pixel in each plane is explicitly encoded into the feature when we concatenated the world coordinates. So \textbf{NNet}($S_i$) is an estimate of the normal at each pixel conditional to the depths in the receptive field of the current slice. For a particular pixel, slices close to the ground truth depth predict good normal estimates, where as slices far from the ground truth predict zero estimates. One way to see this is, if the normal estimate from each slice for a pixel is ${\vec{\mathbf{n_i}}}$, the magnitude of ${\vec{\mathbf{n_i}}}$ can be seen as the correspondence probability at that slice for that pixel. The direction of ${\vec{\mathbf{n_i}}}$ can be seen as the vector aligning with the strong correspondences in the local patch around the pixel in that slice.
\footnote{Refer to the appendix for Visualisation of the NNet slices} 


We train the first module with ground truth depth ($\mathbf{Z_{gt}}$) supervision on $\mathbf{Z_1}$ \& $\mathbf{Z_2}$ along with the ground truth normal (${\vec{\mathbf{n_{gt}}}}$) supervision on (${\vec{\mathbf{n}}}$). The loss function ($\mathcal{L}$) is defined as follows. 
\begin{equation}
\begin{aligned}
    \mathcal{L}_z &= |Z_2 - Z_{gt}|_\mathbf{H} + \lambda_z|Z_1 - Z_{gt}|_\mathbf{H}\\
    \mathcal{L}_n &= |\vec{n} - \vec{n_{gt}}|_\mathbf{H}\\
    \mathcal{L} &= \mathcal{L}_z + \lambda_n\mathcal{L}_n
\end{aligned}
\end{equation} where $|.|_\mathbf{H}$ denotes the Huber norm\footnote{Also referred to as Smooth L1Loss.}. 

\subsection{Depth Normal Consistency}
In addition to estimating depth and normal jointly from the cost volume, we use a novel consistency loss to enforce consistency between the estimated depth and normal maps. 
We utilize the camera model to estimate the spatial gradient of the depth map in the pixel coordinate space using the depth map and normal map. We compute two estimates for $\big(\frac{\partial Z}{\partial u}$, $\frac{\partial Z}{\partial v}\big)$ and enforce them to be consistent.
\begin{figure}
\setlength{\abovecaptionskip}{0pt} 
\setlength{\belowcaptionskip}{0pt}
\begin{center}
\includegraphics[width=\linewidth]{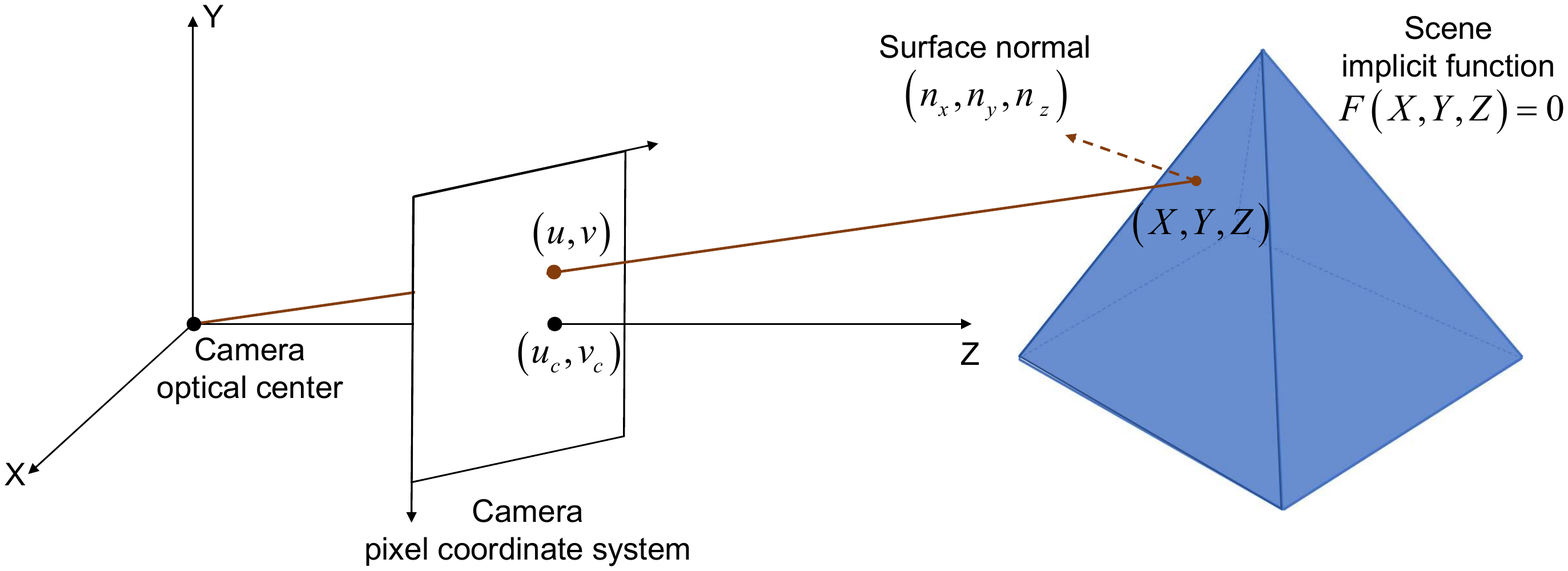}
\end{center}
  \caption{\textbf{Camera Model}
  }
\label{fig:camera_model}
\end{figure}

A pinhole model of the camera is adopted as shown in Figure ~\ref{fig:camera_model}).
\begin{equation}
    \begin{bmatrix}u\\v\\1\end{bmatrix}=\frac1Z\begin{bmatrix}f_x&0&u_c\\0&f_y&v_c\\0&0&1\end{bmatrix}\begin{bmatrix}X\\Y\\Z\end{bmatrix}
\end{equation}
where $(u,v)$ is the corresponding image pixel coordinate of 3D point $(X, Y, Z)$, $(u_c, v_c)$ is the pixel coordinate of camera optical center, and $f_x, f_y$ are the focal lengths for X-axis and Y-axis, respectively.

From the camera model, we can yield:
\begin{equation}\label{eq:2}
\begin{aligned}
X = \frac{Z(u-u_c)}{f_x} &\implies \frac{\partial X}{\partial u} = \frac{u-u_c}{f_x}\frac{\partial Z}{\partial u} + \frac{Z}{f_x}\\
Y = \frac{Z(v-v_c)}{f_y} &\implies \frac{\partial Y}{\partial u} = \frac{v-v_c}{f_y}\frac{\partial Z}{\partial x}
\end{aligned}
\end{equation}

\noindent{Estimate 1:}

The spatial gradient of the depth map can first be computed from the depth map by using a Sobel filter:
\begin{equation}
\begin{aligned}
\bigg(\frac{\partial Z}{\partial u}, \frac{\partial Z}{\partial v}\bigg)_{1} = \bigg(\frac{\Delta Z}{\Delta u}, \frac{\Delta Z}{\Delta v}\bigg)
\end{aligned}
\end{equation}

\noindent{Estimate 2:}

We assume the underlying scene to be of a smooth surface which can be expressed as an implicit function $F(X,Y,Z)=0$. The normal map $\vec{n}$ is an estimate of the gradient of this surface.

\begin{equation}\label{eq:1}
\begin{aligned}
&\vec{n} = (n_x, n_y, n_z) = \bigg(\frac{\partial F}{\partial X}, \frac{\partial F}{\partial Y}, \frac{\partial F}{\partial Z}\bigg)\\
&\implies\frac{\partial Z}{\partial X} = \frac{-n_x}{n_z}, \frac{\partial Z}{\partial Y} = \frac{-n_y}{n_z}
\end{aligned}
\end{equation}

Therefore, we can derive the second estimate of the depth spatial gradient by:

\begin{equation}\label{eq:3}
\begin{aligned}
\bigg(\frac{\partial Z}{\partial u}\bigg)_{2} &= \frac{\partial Z}{\partial X}\frac{\partial X}{\partial u} + \frac{\partial Z}{\partial Y}\frac{\partial Y}{\partial u}\\
&= \frac{\big(\frac{-n_xZ}{n_zf_x}\big)}{1+\big[\frac{n_x(u-u_c)}{n_zf_x}\big] + \big[\frac{n_y(v-v_c)}{n_zf_y}\big]}
\end{aligned}
\end{equation}
%
Similarly,
\begin{equation}\label{eq:4}
    \bigg(\frac{\partial Z}{\partial v}\bigg)_{2} = \frac{\big(\frac{-n_yZ}{n_zf_y}\big)}{1+\big[\frac{n_x(u-u_c)}{n_zf_x}\big] + \big[\frac{n_y(v-v_c)}{n_zf_y}\big]}
\end{equation}

The consistency loss $\mathcal{L}_c$ is given as the Huber norm of the deviation between the two estimates $\big(\frac{\partial Z}{\partial u}$, $\frac{\partial Z}{\partial v}\big)_1$ and $\big(\frac{\partial Z}{\partial u}$, $\frac{\partial Z}{\partial v}\big)_2$ :
\begin{equation}
    \mathcal{L}_c = \bigg|\bigg(\frac{\partial Z}{\partial u}, \frac{\partial Z}{\partial v}\bigg)_{1} - \bigg(\frac{\partial Z}{\partial u}, \frac{\partial Z}{\partial v}\bigg)_{2}\bigg|_\mathbf{H}
\end{equation}

The second estimate of the depth spatial gradient depends only the absolute depth of the pixel in question and not the depths of the neighboring pixels. We obtain the local surface information from the normal map, which we deem more accurate and easier to estimate. Our consistency formulation not only enforces constraints between the relative depths in a pixel's neighborhood but also the absolute depth. Previous approaches like ~\cite{qi2018geonet}, ~\cite{DBLP:conf/cvpr/ZhangF18} enforce consistency between depth and normal maps by constraining the local surface tangent obtained from the depth map to be orthogonal to the estimated normal. These approaches typically enforce constraints on the spatial depth gradient in the world coordinate space, where as we enforce them in the pixel coordinate space. We convert the previous approach into a depth gradient consistency formulation and provide a detailed comparison along with experiments on SUN3D dataset in the appendix.


In our pipeline, we implement this loss in an independent module. We use a UNet~\cite{DBLP:journals/corr/RonnebergerFB15} with the raw depth and normal estimates as inputs to predict the refined depth and normal estimates. We train the entire pipeline in a modular fashion. We initially train the first module with loss $\mathcal{L}$ and then add the second module with the consistency loss $\mathcal{L}_c$ in conjunction with the previous losses. Moreover, our loss function can also be used for any depth refinement/completion method including single-view estimation given an estimate of the normal map.



\section{Experiments}
\subsection{Datasets}
We use SUN3D~\cite{DBLP:conf/iccv/XiaoOT13}, RGBD~\cite{DBLP:conf/iros/SturmEEBC12} and Scenes11~\cite{DBLP:journals/corr/ChangFGHHLSSSSX15} datasets for training our end-to-end pipeline from scratch. The train set contain 166,285 image pairs from 50420 scenes (SUN3D: 29294, RGBD: 3373, Scenes11: 17753). Scenes11 is a synthetic dataset whereas SUN3D and RGBD consist of real word indoor environments. We test on the same split as previous methods and report the common quantitative measures of depth quality: absolute relative error (Abs Rel), absolute relative inverse error (Abs R-Inv), absolute difference error (Abs diff), square relative error (Sq Rel), root mean square error and its log scale (RMSE and RMSE log) and inlier ratios ($\delta < 1.25^i$ where $i \in \{1,2,3\}$). 

We also evaluate our method on a different class of datasets used by other state-of-the-art methods. We train and test on the Scene Flow datasets~\cite{MIFDB16} which consist of 35454 training and 4370 test stereo pairs in 960$\times$540 resolution with both synthetic and natural scenes. The metrics we use on this dataset are the popularly used average End Point Error (EPE) and the 1-pixel threshold error rate.

Further, we evaluate our task on ScanNet~\cite{DBLP:conf/cvpr/DaiCSHFN17}. The dataset consists of 94212 image pairs from 1201 scenes. We use the same test split as in~\cite{DBLP:conf/cvpr/ZengTHYSCW19}. We follow~\cite{yao2018mvsnet} for neighboring view selection, and generate ground-truth normal map following~\cite{fouhey2013data}. 
We use ScanNet to evaluate the performance of the surface normal estimation task too. We use the mean angle error (mean) and median angle error (median) per pixel. In addition, we also report the fraction of pixels with absolute angle difference with ground truth less than t where t $\in$ \{11.25\degree, 22.5\degree, 30\degree \}. For all the tables, we represent if a lower value of a metric is better with ($\downarrow$) and if an upper value of a metric is better with ($\uparrow$).

For all the experiments, to be consistent with other works, we use only two views to train and evaluate. ScanNet~\cite{DBLP:conf/cvpr/DaiCSHFN17} provides depth map and camera pose for each image frame. To make it appropriate for stereo evaluation, view selection is a crucial step. Following Yao~\etal~\cite{yao2018mvsnet}, we calculate a score $s(i, j) = \sum_{\mathbf{p}} \mathcal{G}(\theta_{ij}(\mathbf{p}))$ for each image pair according to the sparse points, where $\mathbf{p}$ is a common track in both view $i$ and $j$, $\theta_{ij}(\mathbf{p}) = (180/\pi)\arccos((\mathbf{c}_i - \mathbf{p}) \cdot (\mathbf{c}_j - \mathbf{p}))$ is $\mathbf{p}$'s baseline angle and $\mathbf{c}$ is the camera center. $\mathcal{G}$ is a piece-wise Gaussian function \cite{zhang2015joint} that favors a certain baseline angle $\theta_0$:
\[ \mathcal{G}(\theta) =  \left\{
\begin{array}{ll}
     \exp(-\frac{(\theta - \theta_0)^2}{2\sigma_1^2}), \theta \leq \theta_0 \\
     \exp(-\frac{(\theta - \theta_0)^2}{2\sigma_2^2}), \theta > \theta_0 \\
\end{array} 
\right. \]

In the experiments, $\theta_0$, $\sigma_1$ and $\sigma_2$ are set to 5$\degree$, 1 and 10 respectively. We generate ground-truth surface normal maps following the procedure of~\cite{fouhey2013data}.

\begin{table*}[t]
\begin{center}
\begin{tabular}{|l|l| p{1.1 cm} p{1.1 cm} p{1.1 cm} p{1.1 cm} p{1.0 cm} c c c|}
\hline
Dataset & Method & Abs Rel($\downarrow$) & Abs diff($\downarrow$) & Sq Rel($\downarrow$) & RMSE\newline($\downarrow$) & RMSE log($\downarrow$) & \pbox{1.1cm}{$\delta<$1.25\newline($\uparrow$)} & \pbox{1.1cm}{$\delta<$1.25\textsuperscript{2}\newline ($\uparrow$)} & \pbox{1.2cm}{$\delta<$1.25\textsuperscript{3}\newline($\uparrow$)} \\
\hline\hline
MVS & COLMAP~\cite{DBLP:conf/cvpr/SchonbergerF16}  &  0.3841 & 0.8430 & 1.257  & 1.4795  & 0.5001  & 0.4819  & 0.6633  & 0.8401 \\
(Outdoor)    & DeMoN~\cite{DBLP:conf/cvpr/UmmenhoferZUMID17}   &  0.3105 & 1.3291 & 19.970 & 2.6065 & 0.2469 & 0.6411 & 0.9017 & 0.9667 \\
    & DeepMVS~\cite{DBLP:conf/cvpr/HuangMKAH18} &  0.2305 & 0.6628 & 0.6151 & 1.1488 & 0.3019 & 0.6737 & 0.8867 & 0.9414\\ 
    & DPSNet-U~\cite{DBLP:conf/iclr/ImJLK19}  &  0.0813 & 0.2006 & 0.0971 & 0.4419 & 0.1595 & 0.8853 & 0.9454 & 0.9735\\
    \cline{2-10}
    & \textbf{Ours}    &  \textbf{0.0679} & \textbf{0.1677} & \textbf{0.0555} & \textbf{0.3752} & \textbf{0.1419} & \textbf{0.9054} & \textbf{0.9644} &  \textbf{0.9879} \\
\hline
SUN3D & COLMAP~\cite{DBLP:conf/cvpr/SchonbergerF16}  &  0.6232 & 1.3267 & 3.2359 & 2.3162 & 0.6612 & 0.3266 & 0.5541 & 0.7180\\
(Indoor)  & DeMoN~\cite{DBLP:conf/cvpr/UmmenhoferZUMID17}   &  0.2137 & 2.1477 & 1.1202 & 2.4212 & 0.2060 & 0.7332 & 0.9219 & 0.9626\\
    & DeepMVS~\cite{DBLP:conf/cvpr/HuangMKAH18} &  0.2816 & 0.6040 & 0.4350 & 0.9436 & 0.3633 & 0.5622 & 0.7388 & 0.8951\\ 
    & DPSNet-U~\cite{DBLP:conf/iclr/ImJLK19} &  0.1469 & 0.3355 & 0.1165 & 0.4489 & 0.1956 & 0.7812 & 0.9260 & 0.9728\\
    \cline{2-10}
    & \textbf{Ours}    &   \textbf{0.1271} & \textbf{0.2879} & \textbf{0.0852} & \textbf{0.3775} & \textbf{0.1703} & \textbf{0.8295} & \textbf{0.9437} & \textbf{0.9776}\\
\hline
RGBD & COLMAP~\cite{DBLP:conf/cvpr/SchonbergerF16}  & 0.5389 & 0.9398 & 1.7608 & 1.5051 & 0.7151 & 0.2749 & 0.5001 & 0.7241\\
(Indoor) & DeMoN~\cite{DBLP:conf/cvpr/UmmenhoferZUMID17}  &  0.1569 & 1.3525 & 0.5238 & 1.7798 & 0.2018 & 0.8011 & 0.9056 & 0.9621\\
    & DeepMVS~\cite{DBLP:conf/cvpr/HuangMKAH18} &  0.2938 & 0.6207 & 0.4297 & 0.8684 & 0.3506 & 0.5493 & 0.8052 & 0.9217\\
    & DPSNet-U~\cite{DBLP:conf/iclr/ImJLK19} &  0.1508 & 0.5312 & 0.2514 & 0.6952 & 0.2421 & 0.8041 & 0.8948 & 0.9268\\
    \cline{2-10}
    & \textbf{Ours} &   \textbf{0.1314} & \textbf{0.4737} & \textbf{0.2126} & \textbf{0.6190} & \textbf{0.2091} & \textbf{0.8565} & \textbf{0.9289} & \textbf{0.9450}\\
\hline
Scenes11 & COLMAP~\cite{DBLP:conf/cvpr/SchonbergerF16}  &  0.6249 & 2.2409 & 3.7148 & 3.6575 & 0.8680 & 0.3897 & 0.5674 & 0.6716\\
(Synthetic)    & DeMoN~\cite{DBLP:conf/cvpr/UmmenhoferZUMID17}   &  0.5560 & 1.9877 & 3.4020 & 2.6034 & 0.3909 & 0.4963 & 0.7258 & 0.8263\\
    & DeepMVS~\cite{DBLP:conf/cvpr/HuangMKAH18} &  0.2100 & 0.5967 & 0.3727 & 0.8909 & 0.2699 & 0.6881 & 0.8940 & 0.9687\\
    & DPSNet-U~\cite{DBLP:conf/iclr/ImJLK19} &  0.0500 & 0.1515 & 0.1108 & 0.4661 & 0.1164 & 0.9614 & 0.9824 & 0.9880\\
    \cline{2-10}
    & \textbf{Ours}  &  \textbf{0.0380} & \textbf{0.1130} & \textbf{0.0666} & \textbf{0.3710} & \textbf{0.0946} & \textbf{0.9754} & \textbf{0.9900} & \textbf{0.9947}\\ 
\hline
\end{tabular}
\end{center}
\caption{Comparative evaluation of our model on SUN3D, RGBD, Scenes11 and MVS datasets. For all the metrics except the inlier ratios, lower the better. We use the perfomance of COLMAP, DeMoN, and DeepMVS reported in~\cite{DBLP:conf/iclr/ImJLK19}.
}
\label{tab: demon}
\end{table*}

\subsection{Implementation details}
We use 64 levels of depth/disparity while building the cost volume. The hyperparameters $\lambda_z$ and $\lambda_n$ are set to 0.7 and 3 respectively. We train the network without the consistency module first for 30 epochs with ADAM optimizer with a learning rate of $2\times10^{-4}$. Further, we finetune the consistency module with the end-to-end pipeline for 10 epochs with a learning rate of $1\times10^{-4}$. The training process takes 5 days and uses 4 NVIDIA GTX 1080Ti GPUs with a batch size of 12. We use a random crop size of (320 $\times$ 240) during training which can be optionally increased in the later epochs by decreasing the batch size.

\subsection{Comparison with state-of-the-art}\label{ssec:sota}
\begin{figure}
\setlength{\abovecaptionskip}{0pt} 
\setlength{\belowcaptionskip}{0pt}
\begin{center}
\includegraphics[width=\linewidth]{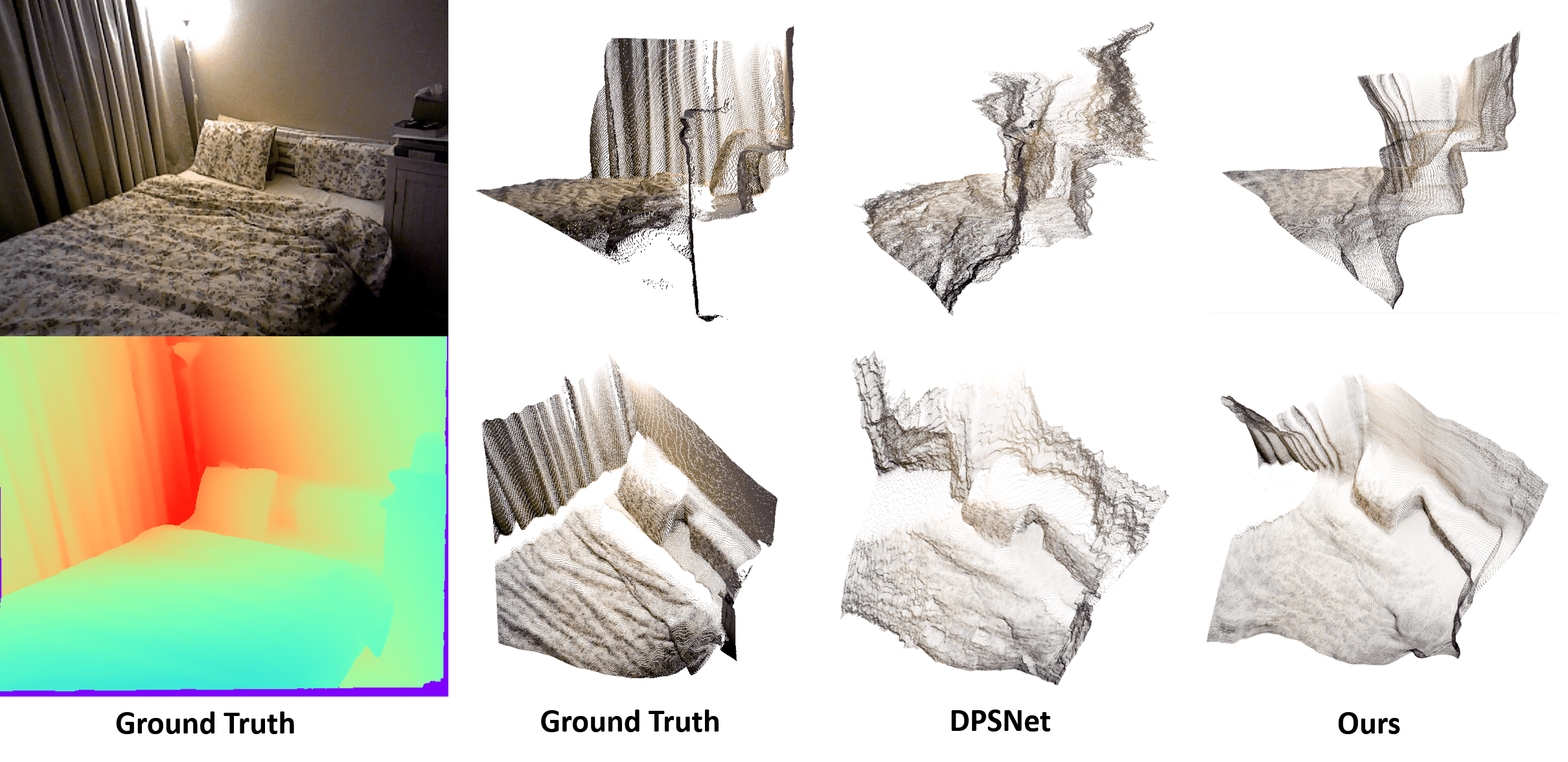}
\end{center}
  \caption{\textbf{Visualizing the depths in 3D for SUN3D.} Two views for the point cloud from depth prediction.
  }
\label{fig:pop}
\end{figure}

For comparisons on the DeMoN datasets (SUN3D, RGBD, Scenes11 and MVS), we choose state-of-the-art approaches of a diverse kind. 
We also evaluate on another dataset MVS~\cite{DBLP:conf/cvpr/SchonbergerF16} containing outdoor scenes of buildings which is not used for training to evaluate generalizability. 
The complete comparison on all the metrics is presented in Table \ref{tab: demon}, and some qualitative results are shown in Fig.~\ref{fig:pop}. Our method outperforms existing methods in terms of all the metrics. Also, our method generates more accurate and smooth point cloud with fine details even in textureless regions (eg. bed, wall).

\begin{table}
\begin{center}
\begin{tabular}{|l|c c|}
\hline
Method & EPE($\downarrow$) & 1-pixel error rate($\downarrow$) \\
\hline\hline
GCNet & 1.80 & 15.6\\
PSMNet & 1.09 & 12.1 \\
DPSNet & 0.80 & 8.4 \\
GANet-15 & 0.84 & 9.9 \\
GANet-deep & 0.78 & 8.7\\
\hline
GANet-NNet & 0.77 & 8.0 \\
Ours & \textbf{0.69} & \textbf{7.0}\\
\hline
\end{tabular}
\end{center}
\caption{Comparative evaluation of our model on Scene Flow datasets. For all the metrics, lower the better.}
\label{tab: sflow}
\end{table}
We compare our performance against similar cost-volume based approaches GCNet~\cite{DBLP:journals/corr/KendallMDHKBB17}, PSMNet~\cite{DBLP:journals/corr/abs-1803-08669}) and GANet~\cite{DBLP:conf/cvpr/ZhangPYT19} which have different choices of cost aggregation. Since we use the same testing protocol, we use the performance of GCNet, PSMNet and GANet-15 as reported in~\cite{DBLP:conf/cvpr/ZhangPYT19}. We obtain the performance of GANet-deep which uses a deeper network with more 3D convolutions from the authors' website. Further, we append our NNet branch to the existing GANet architecture by passing the cost volume of GANet through our NNet and train this branch simultaneously with the full GANet architecture. We call this GANet-NNet. Finally, we also train DPSNet on scene flow datasets to confirm that the better performance is due to normal supervision rather than better cost aggregation or a better architecture. 

\begin{table}[h]
\begin{center}
\small
\begin{tabular}{|p{1.0 cm}|p{1.15 cm}| p{0.8 cm} p{0.8 cm} p{0.8 cm} p{0.8 cm}|}
\hline
 Dataset & Method & Abs Rel($\downarrow$) & Abs diff($\downarrow$) & Sq Rel($\downarrow$) & RMSE\newline($\downarrow$) \\

\hline\hline
ScanNet & DPSNet & 0.1258 & 0.2145 & 0.0663 & 0.3145\\
      & Ours & 0.1150 & 0.2068 & 0.0577 & 0.3009 \\
      & Ours-$\mathcal{L}_c$ & \textbf{0.1070} & \textbf{0.1946} & \textbf{0.0508} & \textbf{0.2807}\\
\hline
SUN3D & DPSNet & 0.1470 & 0.3234 & 0.1071 & 0.4269\\
      & Ours & 0.1332 & 0.3038 & 0.0910 & 0.3994\\
      & Ours-$\mathcal{L}_c$ & \textbf{0.1247} & \textbf{0.2848} & \textbf{0.0791} & \textbf{0.3671}\\
\hline
\end{tabular}
\end{center}
\caption{Comparative evaluation of our consistency loss.}
\label{tab:cons}
\end{table}

We also evaluate the performance of our consistency loss on SUN3D and ScanNet datasets. We train DPSNet only on SUN3D as well as ScanNet independently along with our method and present the results and present them in Table \ref{tab:cons}. We observe that our pipeline achieves significantly better performance on all the metrics on the MVS, SUN3D, RGBD, Scenes11 \& SceneFlow datasets. We find that joint multi-view normal and depth estimation helps improve performance on indoor, outdoor, real and synthetic datasets. We further show that our consistency module significantly improves the performance on top of our existing pipeline. We further evaluate the performance on planar and textureless surfaces and visualise the changes in the cost volume due to the addition of NNet.

\subsection{Surface Normal Estimation} 
Table~\ref{tab:normal} compares our cost volume based surface normal estimation with existing RGB-based, depth-based and RGB-D methods. 
We perform significantly better than the depth completion based method and perform similar to the RGB based method. The RGB-D based method performs the best, because of using the ground truth depth data. 

\begin{figure}[h]
\begin{center}
  \includegraphics[width=0.95\linewidth]{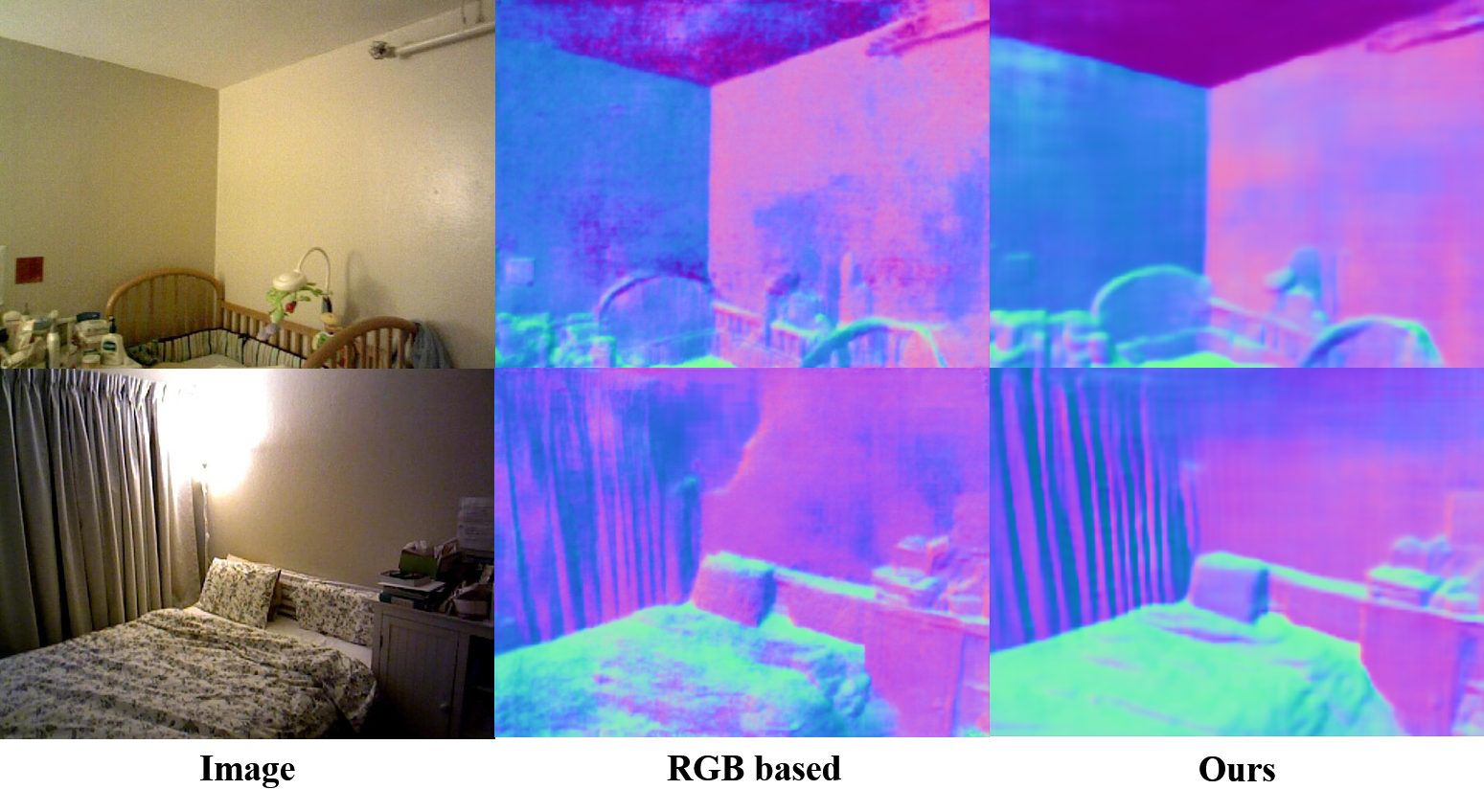}
\end{center}
  \caption{\textbf{Surface Normal Estimation.} Test on SUN3D after training on ScanNet. The RGB-based method is from~\cite{DBLP:conf/cvpr/ZhangSYSLJF17}}
\label{fig:wild}
\end{figure}

\begin{table}[h]
\begin{center}
\begin{tabular}{|p{1.8 cm}| p{0.7 cm} p{0.7 cm} p{0.7 cm} p{0.7 cm} p{0.7 cm}|}
\hline
 Method & Mean\newline($\downarrow$) & Median\newline($\downarrow$) & 11.25$\degree$\newline($\uparrow$) & 22.5$\degree$\newline($\uparrow$) & 30$\degree$\newline($\uparrow$) \\
\hline\hline
RGB-D~\cite{DBLP:conf/cvpr/ZengTHYSCW19}&  14.6 & 7.5 & 65.6 & 81.2 & 86.2 \\
DC~\cite{DBLP:conf/cvpr/ZhangSYSLJF17}  & 30.6 & 20.7 & 39.2 & 55.3 & 60.0\\
\hline
RGB~\cite{DBLP:conf/cvpr/ZhangSYSLJF17} &  31.1 & \textbf{17.2} & \textbf{37.7} & 58.3 & 67.1\\
Ours &  \textbf{23.1} & 18.0 & 31.1 & \textbf{61.8} & \textbf{73.6} \\
\hline
\end{tabular}
\end{center}
\caption{Comparison of normal estimation on ScanNet with single view normal estimation. Note that the RGB-D and depth completion (DC) based methods use ground truth depth. The performances of DC \& RGB-D are from~\cite{DBLP:conf/cvpr/ZengTHYSCW19} and RGB from~\cite{DBLP:conf/cvpr/ZhangSYSLJF17}.}
\label{tab:normal}
\end{table}
We evaluate the surface normal estimation performance in the wild by comparing our method against RGB based methods~\cite{DBLP:conf/cvpr/ZhangSYSLJF17}. We use models trained on ScanNet and test them on images from the SUN3D dataset. We present the results in Table \ref{tab:wild} and visualize a few cases in Fig.~\ref{fig:wild}. We notice that our method generalizes much better in the wild when compared to the single-view RGB based methods. NNet estimates normals accurately not only in regions of low texture but also in regions with high variance in texture (the bed's surface). We attribute this performance to using two views instead of one which reduces the learning complexity of the task and thus generalizes better.

We also observe that irrespective of the dataset, the normal estimation loss as well as the validation accuracies saturate within 3 epochs, showing that the task of normal estimation from cost volume is much easier than depth estimation.
\begin{table}[h]
\begin{center}
\begin{tabular}{|l| p{0.6 cm} p{0.8 cm} p{0.7 cm} p{0.7 cm} p{0.7 cm}|}
\hline
 \pbox{0.9 cm}{\centering Method} & Mean ($\downarrow$) & Median ($\downarrow$) & 11.25$\degree$ ($\uparrow$) & 22.5$\degree$ ($\uparrow$) & 30$\degree$\newline($\uparrow$) \\
\hline\hline
RGB - SUN3D &  31.6 & 25.7 & 17.9 & 45.6 & 57.6\\
Ours - SUN3D & \textbf{22.9} & \textbf{17.0} & \textbf{34.5} & \textbf{63.2} & \textbf{73.6}\\ 
\hline
RGB - MVS &  33.3 & 27.8 & 11.8 & 42.4 & 55.1\\
Ours - MVS & \textbf{27.7} & \textbf{22.4} & \textbf{23.1} & \textbf{52.0} & \textbf{63.9}\\ 
\hline
\end{tabular}
\end{center}
\caption{Generalization performance. Both the models were trained on ScanNet (indoor) and tested on SUN3D (indoor) and MVS (outdoor) datasets}
\label{tab:wild}
\end{table}

\subsection{Visualizing the Cost Volume}
\textbf{Regularization}
Existing stereo methods, both traditional and learning-based ones perform explicit cost aggregation on the cost volume. This is because each pixel has good correspondences only with a few pixels in its neighborhood in the other view. But the cost volume contains many more candidates, to be specific, the number of slices per pixel. Further, in textureless regions, there are no distinctive features and thus all candidates have similar features. This induces a lot of noise in the cost volume also is responsible for to false-positives. 
\begin{figure}
\begin{center}
\includegraphics[width=1\linewidth]{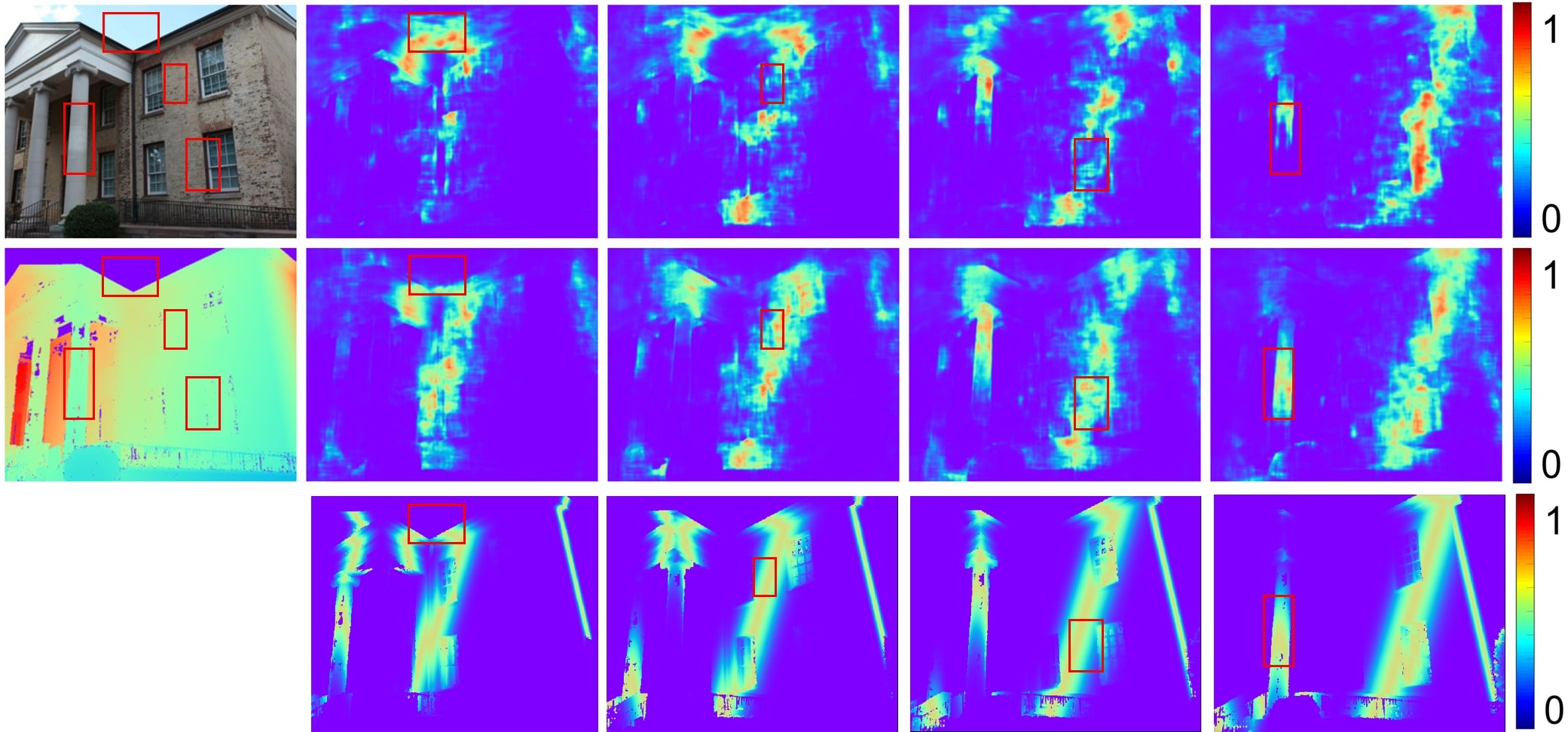}
\end{center}
  \caption{\textbf{Cost slice visualization: } The first column contains the reference image and the ground truth depth map. The first row contains the cost volume slices from DPSNet. The second row contains the same from our network. The third row contains the estimates of ground truth cost slices. This can be seen as a $\mathcal{N}(0,0.01)$ distribution around the ground truth depth corresponding to each slice.}
\label{fig:slices}
\end{figure}
We show that normal supervision during training regularizes the cost volume. Fig.~\ref{fig:slices} visualises the probability volume slices and compares it against those of DPSNet. We consider the un-aggregated probability volume $\mathbf{P_1}$ in both the cases. We visualise the slices at disparities 14, 15, 16 \& 17 (corresponding to depths 2.28, 2.13, 2.0, 1.88) which encompass the wall of the building. The slices of dpsnet are very noisy and do not actually produce good outputs in textureless regions like the walls \& sky and reflective regions like the windows. 

\textbf{Planar and Textureless Regions}
\begin{figure}
\begin{center}
\includegraphics[width=\linewidth]{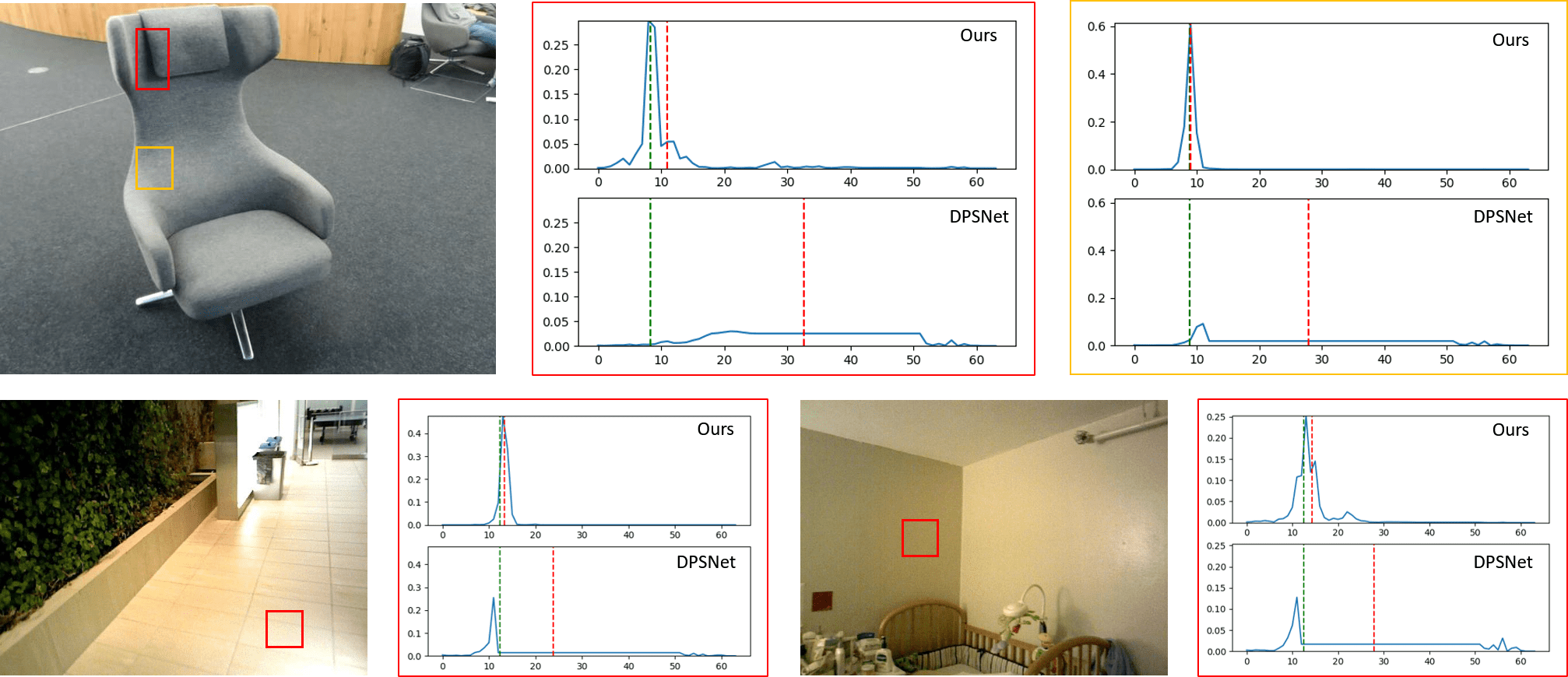}
\end{center}
  \caption{\textbf{Post-softmax probability distributions on disparity}  Green lines illustrate the ground truth disparities while the red lines illustrate the predicted disparities.}
\label{fig:dist}
\end{figure}
We also visualise the softmax distribution at a few regions in Fig.~\ref{fig:dist}. Challenging regions that are planar or textureless or both are chosen. (a) The chair image consists of very less distinctive textures and the local patches on the chair look the same as those on the floor. But given two views, estimation of normal in the regions with curvature is much easier than estimating depth. This fact allows our model to perform better in the these regions(red \& yellow boxes). 
Cost volume based methods that estimate a probability for each of the candidate depths struggle in textureless regions and usually have long tails in the output distribution. Normal supervision provides additional local constraints on the cost volume and suppresses the tails. This further justifies our understanding (from Section \ref{ssec: nnet}) that the correspondence probability is related to the slice's contribution to the normal map. 

We quantify these observations by evaluating the performance of depths $\mathbf{D_1}$ obtained from $\mathbf{P_1}$ against the performance of DPSNet without cost aggregation in Table \ref{tab:no_agg}. 
It shows that normal supervision helps to regularize the cost volume by constraining the cost volume better both qualitatively in challenging cases and quantitatively across the entire test data.

\begin{table}[h]
\begin{center}
\begin{tabular}{|p{1.3 cm}| p{1 cm} p{1 cm} p{1 cm} p{1 cm}|}
\hline
 Method & Abs Rel($\downarrow$) & Abs diff($\downarrow$) & Sq Rel($\downarrow$) & RMSE\newline($\downarrow$) \\

\hline\hline
DPSNet & 0.1274 & 0.3388 & 0.1957 & 0.6230 \\

Ours & \textbf{0.1114} & \textbf{0.3276} & \textbf{0.1466} & \textbf{0.5631}\\
\hline
\end{tabular}
\end{center}
\caption{Test performance without cost aggregation on the DeMoN datasets.}
\label{tab:no_agg}
\end{table}

Further, we quantify the performance of our methods on planar and textureless surfaces by evaluating on semantic classes on ScanNet test images. Specifically we use the eigen13 classes~\cite{DBLP:journals/corr/abs-1301-3572} and report the depth estimation metrics of our methods against DPSNet on the top-2 occuring classes in Table \ref{tab:semantic}. The performance on the remaining classes can be found in the appendix. We show that our methods perform well on all semantic categories and quantitatively show the improvement on planar and textureless surfaces which are usually found on walls and floors. 
\begin{table}[h]
\begin{center}
\begin{tabular}{|p{1 cm} p{1.2 cm}| p{0.8 cm} p{0.8 cm} p{0.8 cm}p{0.9 cm}|}
\hline
 Label & Method & Abs Rel($\downarrow$) & Abs diff($\downarrow$) & Sq Rel($\downarrow$) & RMSE\newline($\downarrow$) \\

\hline\hline
Wall & DPSNet &  0.1340 & 0.2968 & 0.0871 & 0.3599\\
     &   Ours & 0.1255 & 0.2835 & 0.0799 & 0.3436\\
     &   Ours-$\mathcal{L}_c$ &  \textbf{0.1173} & \textbf{0.2690} & \textbf{0.0721} & \textbf{0.3215}\\
\hline
Floor & DPSNet & 0.1116 & 0.2472 & 0.0777 & 0.2973\\
      & Ours & 0.1092 & 0.2242 & 0.0509 & 0.2642\\
      &    Ours-$\mathcal{L}_c$ &   \textbf{0.1037} & \textbf{0.2061} & \textbf{0.0474} & \textbf{0.2561}\\
\hline
\end{tabular}
\end{center}
\caption{Semantic class specific evaluation on ScanNet}
\label{tab:semantic}
\end{table}

\section{Conclusion}
In this paper, we proposed to leverage multi-view normal estimation and apply geometry constraints between surface normal and depth at training time to improve stereo depth estimation. We jointly learn to predict the depth and the normal based on the multi-view cost volume. Moreover, we proposed to refine the depths from depth, normal pairs with an independent consistency module which is trained independently using a novel consistency loss. Experimental results showed that joint learning can improve both the prediction of normal and depth, and the accuracy \& smoothness can be further improved by enforcing the consistency. We achieved state-of-the-art performance on MVS, SUN3D, RGBD, Scenes11 and Scene Flow datasets.


\clearpage

{\begin{flushleft}\LARGE \textbf{Appendix} \end{flushleft}}

\section*{Analysis of $\mathcal{L}_c$}
We first analyse the depth propagation method using normals proposed in ~\cite{qi2018geonet} and reduce it to a form where we can compare it with $\mathcal{L}_c$. In ~\cite{qi2018geonet}, given the depth estimate of pixel i, $Z_i$ is accurate, the depth estimate of neighboring pixel j, $Z_j$ is estimated using, 
\begin{equation}
    Z_j = \frac{n_{x}X_j+n_{y}Y_j+n_{z}Z_j}{(u_i - c_x)n_{x}/f_x+(v_i - c_y)n_{y}/f_y+n_z}
\end{equation}
where $(n_x,n_y,n_z)$ is the normal map estimate at j.
This equation can be rearranged to 
\begin{equation}
\begin{aligned}
    (Z_i - Z_j) &= \frac{Z_iu_i - Z_ju_j}{f_x}\bigg(\frac{-n_x}{n_z}\bigg) + \frac{Z_iv_i - Z_jv_j}{f_y}\bigg(\frac{-n_y}{n_z}\bigg)\\
    (Z_i - Z_j) &= (X_i - X_j)\bigg(\frac{-n_x}{n_z}\bigg) + (Y_i -Y_j)\bigg(\frac{-n_y}{n_z}\bigg)\\
    \Delta Z &= \Delta X\bigg(\frac{-n_x}{n_z}\bigg) + \Delta Y\bigg(\frac{-n_y}{n_z}\bigg)\\
\text{Eq. \ref{eq:1}}\Rightarrow\\ \Delta Z &= \Delta X\bigg(\frac{\partial Z}{\partial X}\bigg) + \Delta Y\bigg(\frac{\partial Z}{\partial Y}\bigg)
\end{aligned}
\end{equation}
From definition of total derivative,
\begin{equation}
    dZ = dX\bigg(\frac{\partial Z}{\partial X}\bigg) + dY\bigg(\frac{\partial Z}{\partial Y}\bigg)
\end{equation}
In ~\cite{qi2018geonet}, the authors use the assumption that neighboring pixels can be assumed to be lying on the same tangent plane, which is the same as approximating \big($\frac{dZ}{dX}$, $\frac{dZ}{dY}$\big) with \big($\frac{\Delta Z}{\Delta X}$, $\frac{\Delta Z}{\Delta Y}$\big). 

We now compare this formulation of depth-normal consistency with ours. Considering neighboring pixels along $X$-direction, $\frac{\Delta Z}{\Delta X} = \frac{\partial Z}{\partial X}$, and similarly, $\frac{\Delta Z}{\Delta Y} = \frac{\partial Z}{\partial Y}$. This formulation can be put as an objective function minimization,
\begin{equation}
    \mathcal{L}_t = \bigg|\bigg(\frac{\Delta Z}{\Delta X}, \frac{\Delta Z}{\Delta Y}\bigg) - \bigg(\frac{\partial Z}{\partial X},\frac{\partial Z}{\partial Y}\bigg)\bigg|_\mathbf{H}
\end{equation}
Our formulation $\mathcal{L}_c$ is,
\begin{equation}
    \mathcal{L}_c = \bigg|\bigg(\frac{\Delta Z}{\Delta u}, \frac{\Delta Z}{\Delta v}\bigg) - \bigg(\frac{\partial Z}{\partial u},\frac{\partial Z}{\partial v}\bigg)\bigg|_\mathbf{H}
\end{equation}

So fundamentally, while previous depth-normal consistencies generally deal in world coordinate space, we concentrate on pixel coordinate space, because the depth map we estimate is a function $Z(u,v)$ in $u,v$. By minimizing $\mathcal{L}_c$, we make the assumption of approximating $\big(\frac{\partial Z}{\partial u}, \frac{\partial Z}{\partial v}\big)$ with $\big(\frac{\Delta Z}{\Delta u}, \frac{\Delta Z}{\Delta v}\big)$, in contrast to approximating 
$\big(\frac{\partial Z}{\partial X}, \frac{\partial Z}{\partial Y}\big)$ with $\big(\frac{\Delta Z}{\Delta X}, \frac{\Delta Z}{\Delta Y}\big)$. The first formulation $\mathcal{L}_t$ enforces depth gradient consistency in world coordinate space with the assumption that the depth gradients are locally linear in world coordinate space. The second formulation $\mathcal{L}_c$ enforces depth gradient consistency in pixel coordinate space with the assumption that the depth gradients are locally linear in pixel coordinate space.

Due to the camera projection geometry, the separation between world coordinates of neighboring pixels in $X$ and $Y$ directions depends on the absolute depth at the pixels. The depth gradient linearity assumption in the world coordinate space, assumes the depth gradient to be locally linear at all depth scales, irrespective of the absolute depth. Where as, in our formulation $\mathcal{L}_c$, the depth gradient in pixel coordinate space $\big(\frac{\partial Z}{\partial u}, \frac{\partial Z}{\partial v}\big)$ depends on the absolute depth of the pixel as shown in equation \ref{eq:3}, \ref{eq:4}. So, our formulation takes into account the scale of separation between points over which the depth gradient is assumed to be linear. 
Furthermore, $\big(\frac{\partial Z}{\partial X}, \frac{\partial Z}{\partial Y}\big) = \big(\frac{-n_x}{n_z},\frac{-n_y}{n_z}\big)$ doesn't depend on the absolute depth value and hence only provides information about the relative depths of the pixels, where as $\big(\frac{\partial Z}{\partial u}, \frac{\partial Z}{\partial v}\big)$ depends on the absolute depth of the pixel locality too.

\space We perform a few experiments to analyse the performance gains due to our novel consistency loss. We freeze the stereo pipeline and train the UNet that takes the raw estimates of depth and normal maps and refines the depth estimate. We train three configurations of it: (1) Pure network based refinement with just ground truth supervision, (2) Simple consistency loss, $\mathcal{L}_t$ as regularizer, (3) Our consistency loss $\mathcal{L}_c$ as regularizer. We analyse the performance of the configurations on the SUN3D dataset which consists of indoor environments with a lot of scope for planar, textureless surfaces and report the results in Table \ref{tab:unet}. Using our consistency loss as regularizer improves the depth estimation accuracy superior to the other approaches.
\begin{table}
\begin{center}
\begin{tabular}{|p{1.3 cm}| p{1 cm} p{1 cm} p{1 cm} p{1 cm}|}
\hline
 Method & Abs Rel($\downarrow$) & Abs diff($\downarrow$) & Sq Rel($\downarrow$) & RMSE\newline($\downarrow$) \\

\hline\hline
Raw & 0.1332 & 0.3038 & 0.0910 & 0.3994\\
UNet & 0.1307 & 0.2863 & 0.0878 & 0.3720 \\
UNet-$\mathcal{L}_t$ & 0.1288 & 0.2980 & 0.0842 & 0.3820 \\
UNet-$\mathcal{L}_c$ & \textbf{0.1247} & \textbf{0.2848} & \textbf{0.0791} & \textbf{0.3671}\\
\hline
\end{tabular}
\end{center}
\caption{Ablation Study of Consistency Loss on SUN3D}
\label{tab:unet}
\end{table}

\section*{Visualization of NNet slices}
\begin{figure}[t]
\begin{center}
  \includegraphics[width=0.8\linewidth]{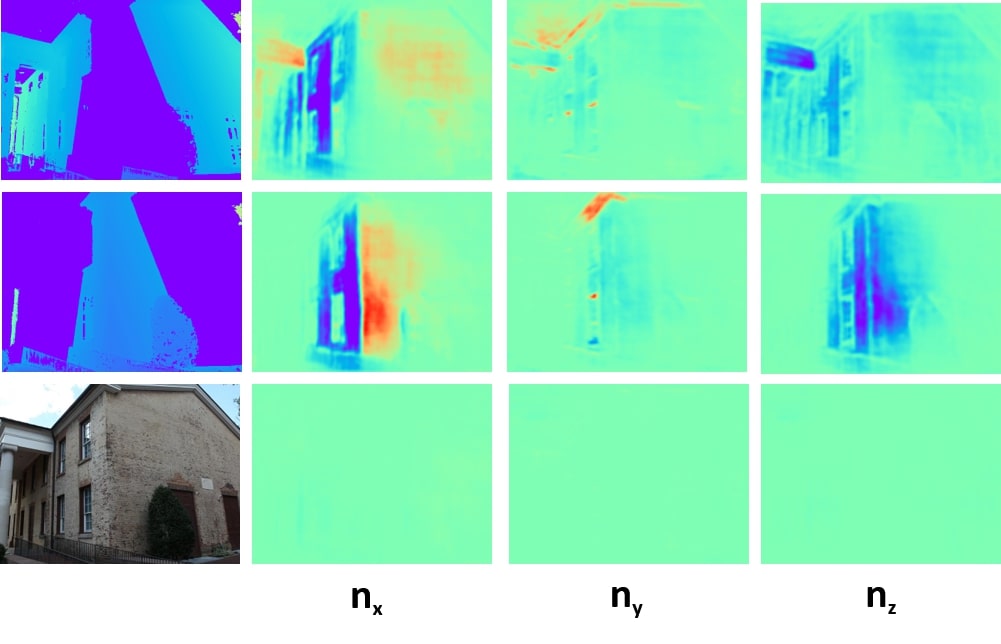}
\end{center}
  \caption{\textbf{Normal Estimation contribution from different slices. }
  The top two rows shows the mask of receptive field and contribution of normal prediction of two slices $S_i$ close to the ground truth depth. The third row shows the sum of the outputs of NNet on all other slices.
  }
\label{fig:nnet-slices}
\end{figure}

We justify the intuition in Section 3.2 in the main paper by visualising the normal estimate contribution from each slice \ie \textbf{NNet}($S_i$) in Figure~\ref{fig:nnet-slices}. The slices in the figure clearly show that only slices with good correspondence probabilities contribute to the output of NNet.

\section*{Semantic class specific evaluation on ScanNet}
We quantify the performance of our methods on planar and textureless surfaces by evaluating on semantic classes on ScanNet test images. Specifically we use the eigen13 classes~\cite{DBLP:journals/corr/abs-1301-3572} and report the depth estimation metrics of our methods against DPSNet. We present the other frequently occuring classes not presented in the paper here in Table \ref{tab:semantic-full}. We show that our methods perform well on all semantic categories and quantitatively show the improvement on planar and textureless surfaces as well which are usually found on walls, floors and ceiling.
\begin{table}[h]
\begin{center}
\begin{tabular}{|p{1 cm} p{1.2 cm}| p{0.8 cm} p{0.8 cm} p{0.8 cm}p{0.9 cm}|}
\hline
 Label & Method & Abs Rel($\downarrow$) & Abs diff($\downarrow$) & Sq Rel($\downarrow$) & RMSE\newline($\downarrow$) \\

\hline\hline
Bed & DPSNet & 0.1291 & 0.1572 & 0.050 & 0.1986 \\
    & Ours & 0.1142 & 0.1449 & 0.0405 & 0.1830\\
    & Ours-$\mathcal{L}_c$ & \textbf{0.1049} & \textbf{0.1347} & \textbf{0.0345} & \textbf{0.1665} \\
\hline
Books & DPSNet & 0.1087 & 0.2281 & 0.0733 & 0.2527\\ 
& Ours & 0.0970 & 0.2176 & 0.0650 & 0.2404\\ 
 & Ours-$\mathcal{L}_c$ & \textbf{0.0942}& \textbf{0.2139} & \textbf{0.0628} & \textbf{0.2334} \\
 \hline
Ceiling & DPSNet & 0.1693 & 0.3429 & 0.1029 & 0.3895 \\
      & Ours & 0.1496 & 0.3189 & 0.0840 & 0.3528 \\
      & Ours-$\mathcal{L}_c$ & \textbf{0.1360} &  \textbf{0.2244} &  \textbf{0.0643} &  \textbf{0.2900} \\
\hline
Chair & DPSNet & 0.1602 &  0.2469 &  0.0836 &  0.3187 \\
      & Ours & 0.1417 &  0.2351 &  0.0697 &  0.3050 \\
      &    Ours-$\mathcal{L}_c$   & \textbf{0.1360} &  \textbf{0.2244} &  \textbf{0.0643} &  \textbf{0.2900} \\
\hline
Floor & DPSNet & 0.1116 & 0.2472 & 0.0777 & 0.2973\\
      & Ours & 0.1092 & 0.2242 & 0.0509 & 0.2642\\
      &    Ours-$\mathcal{L}_c$ &   \textbf{0.1037} & \textbf{0.2061} & \textbf{0.0474} & \textbf{0.2561}\\
\hline
Objects & DPSNet & 0.1305 & 0.2375 & 0.0785 & 0.2934\\
      & Ours & 0.1165 & 0.2237 & 0.0661 & 0.2771\\
      &    Ours-$\mathcal{L}_c$ &   \textbf{0.1095} & \textbf{0.2113} & \textbf{0.0589} & \textbf{0.2587}\\
\hline
Picture & DPSNet & 0.1160 &  0.2991 &  0.0949 &  0.3249\\
& Ours & 0.1110 &  0.2913 &  0.0912 &  0.3167\\
& Ours-$\mathcal{L}_c$ & \textbf{0.1017} &  \textbf{0.2724} &  \textbf{0.0808} &  \textbf{0.2923}\\
\hline
Table & DPSNet & 0.1374 & 0.2211 & 0.0745 & 0.2808 \\
& Ours & 0.1238 & 0.2116 & 0.0646 & 0.2694 \\
& Ours-$\mathcal{L}_c$ &\textbf{0.1164} & \textbf{0.2014} & \textbf{0.0590} & \textbf{0.2545} \\
\hline
Wall & DPSNet &  0.1340 & 0.2968 & 0.0871 & 0.3599\\
     &   Ours & 0.1255 & 0.2835 & 0.0799 & 0.3436\\
     &   Ours-$\mathcal{L}_c$ &  \textbf{0.1173} & \textbf{0.2690} & \textbf{0.0721} & \textbf{0.3215}\\
\hline
Window & DPSNet & 0.1559 & 0.3836 & 0.1353 & 0.4384\\
& Ours & 0.1468 &  0.3605 &  0.1111 &  0.4163 \\
& Ours-$\mathcal{L}_c$ & \textbf{0.1373} &  \textbf{0.3385} &  \textbf{0.1079} &  \textbf{0.3848} \\
\hline
\end{tabular}
\end{center}
\caption{Semantic class specific evaluation on ScanNet. ``DPSNet'' corresponds to the predictions from DPSNet. ``Ours'' corresponds to our predictions before refinement by the consistency module. ``Ours-$\mathcal{L}_c$'' refers to our final predictions}
\label{tab:semantic-full}
\end{table}

\section*{KITTI 2015 Benchmark}
We try to evaluate our method on the KITTI 2015 stereo benchmark. We pre-train our network on the Scene Flow datasets and finetune it on KITTI 2015 train data. We also pre-train GANet-NNet (defined in 4.2 in main paper) on Scene Flow datasets. For GANet, we use the pretrained models the authors provide. We test the performance of these pretrained models first on the KITTI train data without training on it and the report the EPE and 3 pixel error rate in Table \ref{tab:kitti-pretrain}. We then proceed to train on the KITTI 2015 train data and provide the results of the benchmark in Table \ref{tab:kitti}

We observe that the pretrained models generalize better than other methods on KITTI 2015. We obtain significant improvement over DPSNet on the KITTI 2015 test set by adding normal supervision. The KITTI 2015 dataset contains only 200 training images with sparse ground truths with the sparsity increasing as we move to the background. Our ground truth normals are generated using a least squares optimization on the ground truth depths. Sparsity in the ground truth depths makes the generation of very accurate ground truth normals difficult. We see this as a significant problem and affects our performance on KITTI 2015. Despite this problem, GANet-NNet performs better than GANet on the foreground regions.
\begin{table}
\begin{center}
\begin{tabular}{|l|c c|}
\hline
Method & EPE($\downarrow$) & 3-pixel error rate($\downarrow$) \\
\hline\hline
GANet-deep & 1.66 & 10.5\\
\hline
GANet-NNet & 1.64 & 9.7 \\
Ours & \textbf{1.64} & \textbf{8.2}\\
\hline
\end{tabular}
\end{center}
\caption{Evaluation of Scene Flow pretrained models on KITTI2015. For all the metrics, lower the better.}
\label{tab:kitti-pretrain}
\end{table}

\begin{table}
\begin{center}
\begin{tabular}{|l|p{1 cm} p{1 cm} p{1 cm} p{1 cm}|}
\hline
Method & fg-noc($\downarrow$) & both-noc($\downarrow$) & fg-all($\downarrow$) & both-all($\downarrow$) \\
\hline\hline
DPSNet & 6.08 & 4.00 & 7.58 & 4.77\\
GANet-deep &  3.11 & \textbf{1.63} & 3.46 & \textbf{1.81}\\
\hline
GANet-NNet & \textbf{3.04} & 1.70 & \textbf{3.34} & 1.91 \\
Ours & 4.06 & 2.08 & 4.41 & 2.27\\
\hline
\end{tabular}
\end{center}
\caption{Comparative evaluation of our model on KITTI 2015 dataset. For all the metrics, lower the better. \textbf{fg:} Foreground, \textbf{both:} Foreground and Background, \textbf{noc:} Non occluded Pixels, \textbf{all:} All Pixels }
\label{tab:kitti}
\end{table}

\section*{More Qualitative Results}

\begin{figure*}[ht]
\setlength{\abovecaptionskip}{0pt} 
\begin{center}
\includegraphics[width=1\linewidth]{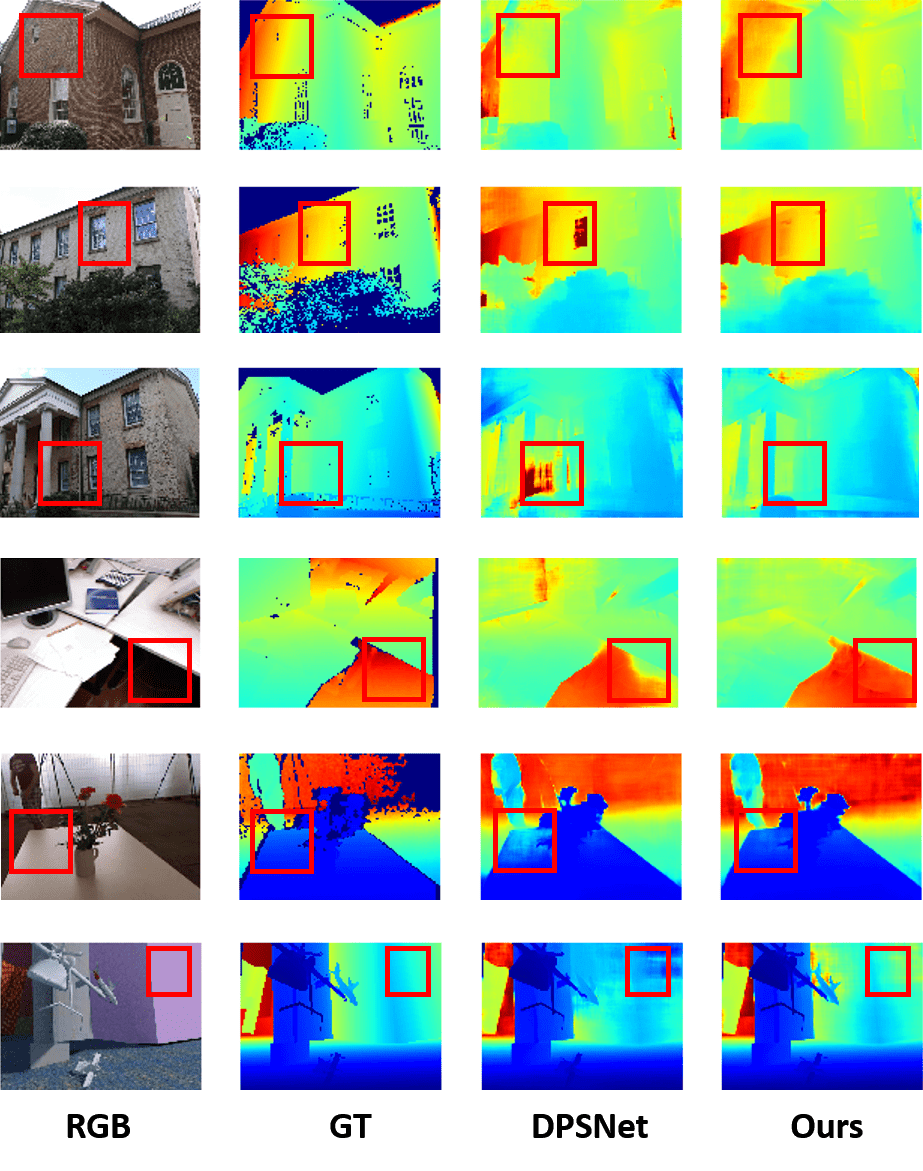}
\end{center}
  \caption{Qualitative comparison of the predicted depth maps. GT represents Ground Truth Depth.}
\label{fig:qual}
\end{figure*}

We present more qualitative results on depth map estimation in Figure \ref{fig:qual}. The examples depict various situations like planar surfaces, reflective surfaces, planar-textureless surfaces and in general overall quality of the prediction. The red boxes on the images illustrate these regions. Our method produces more accurate depth maps when compared to the previous state-of-the-art.

{\small
\bibliographystyle{ieee_fullname}
\bibliography{egbib}
}
\clearpage

\end{document}